\documentclass[11pt]{article}
\usepackage{colacl}
\title{Towards Automated Generation of Scripted Dialogue: \\ Some Time-Honoured Strategies}
\author{Paul Piwek \and Kees van Deemter \\
ITRI -- University of Brighton \\
Watts Building, Moulsecoomb \\ Brighton BN2 4GJ \\
UK \\
Email: Paul.Piwek@itri.bton.ac.uk \hspace*{0.5cm} Kees.van.Deemter@itri.bton.ac.uk }

\begin{document}
\maketitle
\begin{abstract}
The main aim of this paper is to introduce automated generation of scripted dialogue as a worthwhile topic of investigation. In particular the fact that scripted dialogue involves two layers of communication, i.e., uni-directional communication between the author and the audience of a scripted dialogue and bi-directional pretended communication between the characters featuring in the dialogue, is argued to raise some interesting issues. Our hope is that the combined study of the two layers will forge links between research in text generation and dialogue processing. The paper presents a first attempt at creating such links by studying three types of strategies for the automated generation of scripted dialogue. The strategies are derived from examples of human-authored and naturally occurring dialogue. 
\end{abstract}

\section{Introduction}

By a scripted dialogue we mean a
dialogue which is performed by two or more agents on the basis of a
description of that dialogue. This description, i.e., the script,
specifies the actions which are performed in the course of the dialogue
and their temporal ordering. We assume that the script is created in
advance by an author. Automated generation of scripted dialogue
involves a computer programme in the role of the author and execution
of the script by software agents. Andr\'e et al. (2000) coin the term
`presentation team' for such a collection of software agents. In their
words, presentation teams `[...] rather than addressing the user
directly--convey information in the style of performances to be
observed by the user' (Andr\'e et al., 2000:220). 

The plan of this paper is to first motivate
why the study of scripted dialogues is interesting and useful, whilst
also pointing out the limitations and complications of scripted
dialogue. We then present a number of strategies for the automated
generation of scripted dialogue. Our discussion is illustrated by means
of some extracts from mainly scripted dialogues. The paper concludes with
a brief overview of the ongoing {\sc neca} project in which scripted dialogues 
are presented by embodied conversational agents. 

\section{Prospects and Problems}

Scripted dialogues have interesting features, both from a 
theoretical and a practical point of view. Let us start by highlighting A theoretical issue. Scripted dialogues involves two layers of communication. First, in a scripted dialogue there is a layer at which the 
participants of the dialogue {\it mimic} communication with 
each other. The communication is not real because the actions 
of the participants are based on a script; the participants do 
not interpret the actions of the other participants in order to 
determine their own actions. Second, there is a layer of uni-directional 
communication from the author of the script to the audience of the 
dialogue. At this level, a scripted dialogue is very much like a 
monologue. Thus scripted dialogue presents a challenge because it 
requires simultaneous generation of a layer of real and one of 
pretended communication, each layer having its own participants 
with their (real or pretented) goals, beliefs, desires,
 personalities, etc.  

Also from a practical point of view scripted dialogue has something to offer.
Before we go into its advantages, let us, however, first discuss a feature of scripted dialogue which might be perceived as one of its limitations. There is a class of applications which requires the generation of dialogue 
about a subject matter that evolves in real-time. For such applications scripted dialogue is not possible. For instance, 
Andr\'e et al. (2000)'s dialogues between two reporters about a 
live transmission of a {\sc Robocup} soccer event cannot be implemented 
as scripted dialogue: the verbal reports of the agents are determined 
to a large extent by events which evolve in real-time. Hence it is 
impossible to script the verbal reports in advance. Similarly, 
automatically generated scripted dialogues do not lend themselves 
well to user involvement in the dialogue: because the script is 
created in advance there is no scope for reaction to the user's 
contributions.     

These limitations are, however, offset by a large number of benefits. Firstly, it should be noted that for large scale applications it has been argued that a combination of scripted and autonomous behaviours is required. For instance, in the {\sc mre} project at {\sc usc}\footnote{See, e.g., Rickel et al. (2002).} a combination of autonomous and scripted virtual humans is used to create a realistic training environment. Thus applications do not necessarilly force a strict choice between autonomous and scripted behaviour (more specifically dialogue).

Secondly, staying within the education/training domain it should be noted that in the literature on Intelligent Tutoring Systems ({\sc its}) a case has been made for so-called vicarious learning (e.g., Lee et al., 1998; Cox et al., 1999): learning by watching dialogues of other people being taught or engaged in a learning process. Various studies have been carried out in this area and some positive effects of vicarious learning by overhearing dialogue (as opposed to monologue) have been found (see Scott et al., 2000). 

Thirdly, there is a large class of obvious applications for scripted dialogues. The 
dialogue in film scenes, commercials, plays, product demonstrations, 
etc. can be treated as scripted dialogue. These are examples of 
situations in which real-time interaction with the environment or 
a user are not required. 

Finally, there is not only a wide range of potential applications for scripted dialogue, but applying scripted dialogues also has some distinct advantages over relying on dialogue generated by autonomous agents:

\begin{enumerate}
\item The generation of a scripted dialogue requires no potentially complicated and error-prone interpretation of the dialogue acts produced by other autonomous agents. 
\item More time is available for the generation process, because scripted dialogue is not generated in
real time.
\item  Not only more time but also more information is 
available to the generation process of scripted dialogue. 
Whereas in spontaneous dialogue an individual action can only 
be constructed using information about the actions which 
temporally precede it, in scripted dialogue an action can be 
tailored to both the actions which precede and those which
follow it. 
\item It is much easier to create dialogues with 
certain global properties (e.g., a certain pattern of 
turn-taking), because the dialogue is constructed 
by a single author. In spontaneous dialogue, such properties emerge out 
of the autonomous actions of the participants, which makes it
difficult to control them directly. 
\end{enumerate}

\section{Strategies for Scripted Dialogue Generation}

We have already pointed out that in one important respect
scripted dialogue resembles monologue: information 
flows from a single author to an audience. Hovy (1988) was 
one the first to systematically consider how the (communicative) 
goals of an author can be related to various strategies for 
communicating information through a monologue. In particular, 
he describes a natural language generation ({\sc nlg}) 
system ({\sc Pauline}) which implements a number of such 
strategies. He thereby abandoned an assumption which 
was and still is implicit in many {\sc nlg} systems,
namely that a text is generated from a database 
of facts and that the task is mainly one of mapping these facts 
onto declarative sentences which express them. Hovy's work on the influence of pragmatic factors on natural language generation is currently followed up by various researchers involved in building embodied conversational agents (for a bibliography of recent work in this newly emerging area of natural language generation see Piwek, 2002).

The new picture that emerges is one where an author uses 
a text as a device for influencing the attitudes of his or her 
audience. Amongst the attitudes which an author might want to 
influence are the beliefs, intentions (plans for action), goals, 
desires and opinions (judgements about whether something is good, 
bad or neutral) of the audience. All of the aforementioned 
attitudes are {\em about} something (that is, they are intentional; 
see, for instance, Searle, 1983). 
Roughly speaking, one can discern attitudes which are about the 
subject matter/topic of a text and those which pertain to the 
context (e.g., the author, the audience and the 
relation between the two). Attitudes about 
the context are normally communicated implicitly. Hovy discusses 
how style (formal, informal, forceful) can be used to do so. Generally speaking, everything that 
is discussed explicitly in a text is part of its subject matter. Thus, whenever contextual aspects are discussed explicitly (e.g., `I am 
your boss, therefore listen to what I have to say'), they become also 
part of the subject matter. 
This leaves us with a class of information that is neither discussed explicitly (and therefore not part of the subject matter) and is neither part of the context. For instance, take an opinion about a person which can be expressed explicitly  as in `X is a bad guy', but also implicitly as in `X killed John'.\footnote{Hovy (1988) discusses how reporting that a person is the actor of an action which is generally considered to be bad can be used to implicitly convey the opinion that the person is bad.} 

Scripted dialogue offers the same communicative opportunities
(i.e., for communicating facts and influencing an audience in other
ways)
as ordinary text, plus a number of other ones in addition. To
illustrate some of the issues, let us summarize one of the first 
dialogues written by the humanist philosopher Erasmus of Rotterdam
in 1522 (Erasmus, 1522).\footnote{Large parts of the dialogue were omitted,
reworded, or summarized, since we wil be focusing on a specific set
of issues. The (very tentative) translation is our own.}

{\small
\begin{tabbing}

xxX \= xxx \= xxx \= \kill

E1 \> 1. \> A: \> Where have you been?\\
\> 2. \>  C: \> I was off to Jerusalem on \\
\> \> \> pilgrimage.\\
\> 3. \> A: \> Why?\\
\> 4. \> C: \> Why do others go?\\
\> 5. \> A: \> Out of folly if I'm not mistaken.\\
\> 6. \> C: \> That's right; glad I'm not the \\
\> \> \> only one though.\\
\> 7. \> A: \> Was the trip worthwhile?\\
\> 8. \> C: \> No.\\
\> 9. \> A: \> What did you see? \\
\> 10 \> C: \> Pilgrims causing mayhem.\\
\> 11. \> A: \> Were you morally uplifted?\\
\> 13 \> C: \> No, not at all.\\
\> 14 \> A: \> Did you get richer?\\
\> 15. \> C: \> No, quite the contrary. \\
\> 16 \> A: \> Was there nothing good about \\
\>  \> \> the trip then?\\
\> 17 \> C: \> Yes, in fact there was. In \\ 
\> \> \> particular, I can  now entertain \\
\> \> \> others with my lies, \\
\> \> \> like other pilgrims do.\\
\> 18. \> A: \> But that's not very decent, \\
\> \> \> is it? [...]\\
\> 19. \> C: \> True. But I may also be able to \\
\> \> \> talk others out of the idea of \\
\> \> \> pilgrimage.\\
\> 20. \> A: \> I wish you had talked {\em me} out \\
\> \> \> of it.\\
\> 21. \> C: \> What? Have you been as stupid \\
\> \> \> as I?\\
\> 22. \> A: \> I've been to Rome and \\
\> \> \> Santiago de Compostela.\\
\> 23. \> C: \> Why? \\
\> 24. \> A: \> Out of folly I guess [because ...]\\
\> 25 \> C: \> So why did you do it?\\
\> 26. \> A: \> My friends and I vowed to go \\
\> \> \>  when we were drunk.\\
\> 27. \> C: \> Surely a decision worth taking \\
\> \> \> when you're drunk [...] \\
\> \> \>Did everyone arrive back home \\
\> \> \> safely? \\
\> 28. \> A: \> All except three: Two died; \\
\> \> \> the third we left but he's \\ 
\> \> \>   probably in heaven now.\\
\> 29. \> C: \> Why? Was he so pious? \\
\> 30. \> A: \> No, he was a scoundrel.\\
\> 31. \> C: \> Then why is he in heaven? \\
\> 32. \> A: \> Because he had plenty of \\
\>  \> \> letters of indulgence with \\
\>  \> \> him [...]\\
\> 33. \> A: \> Don't get me wrong: I'm \\
\>  \> \> not against letters of indulgence \\
\>  \> \> but I have more admiration for \\ 
\> \> \>  someone who leads a virtuous \\
\> \> \>  life. Incidentally, when do we \\
\> \> \>  go to these parties that you \\
\> \> \>  mentioned?\\
\> 34. \> C: \> Let's go as soon as we can, \\
\> \> \> and add to other pilgrims' \\
\> \> \> lies.
\end{tabbing}
}

\noindent 
The central question that we will start addressing in this 
paper is `what strategies for influencing the attitudes of the 
audience are specific to scripted dialogue?' A number of these 
strategies will be introduced below. We will return to Erasmus' dialogue at various points in our discussion and highlight parts of the dialogue which illustrate the aforementioned strategies.   

\subsection{Strategies of information distribution} 

As we have seen, the main difference between monologue and 
scripted dialogue is 
that the latter communicates with the audience {\em via}
the (pretended) communication between the dialogue participants.
This has a number of immediate effects:

\begin{enumerate}

\item The author can let a participant say something without
being directly responsible for the content.
\item Each participant can represent a particular chunk of 
information, making the combined content more easily digestible.
\item In particular, the participants can represent different points 
of view on the same subject matter, which may even be inconsistent
with each other.
\item One participant may express an opinion concerning 
something the other participant
has raised. 

\end{enumerate}

\noindent
{\em Point 1} was clearly relevant to Erasmus, in whose case direct
criticism of the Catholic church could have made him a target for
the Inquisition. (A's last utterance seems intended to further
milden any criticism.)
{\em Point 2} is relevant because each of the two
participants represents a particular journey. Both journeys could
have been related in one monologue, but this might have led to
confusion and would certainly have been less exciting to read.
In fact, it is patently clear from reading the whole
dialogue that one participant's questions are used for
making more lively what would otherwise have been a story with some rather boring parts. 

{\em Point 3} is not directly relevant in the case of 
the present dialogue but the very fact that both participants 
agree on all essentials can only reinforce the strengths of 
Erasmus' implied position. (See also under Strategies of 
Emphasis.) {\em Point 4} is relevant again, since
both participants frequently express evaluative opinions
concerning various elements of the two stories. In many
cases, evaluative opinions are expressed in highly indirect
fasion, and this brings us to a second class of strategies.  

\subsection{Strategies of association} 

One way to influence the attitude of the audience about, for instance, 
a person is to mention the person in combination with something else 
to which the audience already has the intended attitude. Hovy (1988)
suggests, for example, that we can make somebody look good, bad or 
neutral by presenting him or her as the actor of an action which is 
generally (or specifically by the audience) perceived to be good, 
bad or neutral, respectively: ``Mike killed Jim'' makes Mike 
look bad, whereas ``Mike rescued Jim'' makes him look good.

In fact, this strategy seems to be an instance of a more generally applicable strategy: {\it To convey that X has property P, one can present X in combination with something which has or implies property P}. Thus, to convey that Mike is a clever guy we might say ``Mike managed to solve this partial differential equasion in no time'', i.e., Mike is presented as being able to solve a difficult problem quickly, which implies being clever. 

Information conveyed in a text is not only presented in the context of other information which can influence its interpretation but also {\it by} the author (and possibly speaker) of the text. If any 
properties of this author/speaker are known, these can rub off on the 
points s/he is trying to make. The appeal to this tendency in an argument is considered to be a fallacy (Argumentum ad Hominem).  For instance, one might argue that 
Bacon's philosophy is untrustworthy because he was removed from his 
chancellorship for dishonesty (Copi, 1972:72). 

Scripted dialogue lends it particularly well to this type of
association.
 The presence of the second layer of communication allows the author to 
 distribute communicative acts over characters which were conceived 
 by the author. These characters can be given certain 
 traits which influence the interpretation by the audience of what 
 they say in the dialogue. These traits can be 
 conveyed by various means. In the case of Erasmus'
dialogue, the fact that the protagonists and their pilgrim friends
are avid partygoers -- evidently something Erasmus didn't approve 
of -- is used to discredit their pilgrimage. If the dialogue is enacted by a collection 
 of embodied agents, their physical appearance can be used. 
 Alternatively, certain characteristics of the dialogue can also 
 suggest a particular property.  For instance, Thomas (1989) 
 discusses various ways in which a speaker can come across as 
 dominant or an authority (interruptions, abrupt changes of 
 topic, marking new stages in the interaction, metadiscoursal 
 comments, etc.). The following is an example of a marking of a 
 new stage in the interaction taken from Thomas (1989:146):
\\
\\
{\small
\begin{tabular}[t]{l l p{5.8cm} }
E2 & A: & Okay that's that part. The next part what I want to deal with is 
  your suitability to remain as a CID officer. 
\end{tabular}
}

\subsection{Strategies of emphasis} \label{emphasis}

For various reasons, an author might want to highlight certain 
information and suppress other information. In a monologue, 
repetition of information signals emphasis. For instance, 
de Rosis and Grasso (2000) analyse an explanation text about drug 
prescription and point out that certain information is rather redundant,
i.e., repeated with identical or equivalent wording such as:
\\
\\
{\small
\begin{tabular}[t]{l p{6.8cm} }
E3 & ``The good is news is that we do have tablets that are very effective
for treating TB'', and ``but it is something we can do something about''.
\end{tabular}
}
\\
\\
\noindent 
Here it seems that positive information is repeated intentionally. 
In dialogue, repetition can be achieved naturally due to the presence 
of two interlocutors at the second layer of communication. Consider 
the dialogue fragment below from Twain (1917:11) between a young man and an old man. 

{\small
\begin{tabbing}
xxxx \= xxxx \= xxxx \= \kill
E4 \>1. \> {\sc y.m.} \> What detail is that? \\
\>2. \> {\sc  o.m.} \> The impulse which moves a \\
\> \> \> person to do things -- the \\
\> \> \> only impulse that ever moves \\
\> \> \> a person to do a thing. \\
\>3. \> {\sc y.m.} \> The {\em only} one! Is there \\
\> \> \>but one? \\
\>4. \> {\sc  o.m.} \> That is all. There is only \\
\> \> \> one. \\
\>5. \> {\sc  y.m.} \> Well, certainly that is a \\
\> \> \> strange enough doctrine. \\
\> \> \>What is the sole impulse that \\
\> \> \> ever moves a person to \\
\> \> \> do a thing? \\
\>6. \> {\sc  o.m.} \> The impulse to {\em content} \\
\> \> \> {\em his own spirit}--the {\em necessity} \\
\> \> \> of contending his own spirit \\ 
\> \> \> and {\em winning its approval}.        
\end{tabbing}
}

\noindent
Here turns 3. and 4., which form a subdialogue, are both about 
the claim that there is {\em only one} impulse which moves a person to
do 
things. Now imagine that we have an algorithm which has already distributed the information which it wants to get across to the audience amongst the dialogue participants. Let us call this step I. During step II, the algorithm will determine how this information can be conveyed through a sequence of turns. To generate E4, we might at this stage produce the sequence: 1., 2., 5., 6. Step III would involve the addition of further turns for the purpose of emphasizing information. During step III, such an algorithm could insert subdialogues like the one above (3., 4.), if there are any matters which need particular emphasis.

Note that Erasmus also employs this strategy. For instance, in Erasmus' Dialogue (E1), the turns 4. and 5. form a subdialogue which a system like the one proposed here could insert during step III, after already having created the sequence 1., 2., 3., 6., ... 

Note that the sketched approach presupposes that realization (both verbal and non-verbal) is performed only after steps I -- III; during steps I -- III the algorithm manipulates abstract descriptions of the semantic and pragmatic content of the utterances. The reason for this is that before step III, the algorithm can not yet know whether to realize the beginning of turn 6. as `That's right' or `Out of folly', since this depends on whether the subdialogue (4., 5.) is inserted or not. 

%

\section{Application in the NECA project}

The work reported in this paper is carried out in the context of the {\sc neca} project which started in October 2001 and has a duration of 2.5 years.\footnote{{\sc neca} stands for Net Environment for Embodied Emotional Conversational Agents. The project is funded by the {\sc Ec}. The partners in the project are: {\sc Dfki}, {\sc Ipus} (University of the Saarland), {\sc Itri} (University of Brighton), {\sc \"ofai}, Freeserve and Sysis {\sc ag}. Further details can be found at http://www.ai.univie.ac.at/NECA/.} In this project a system is being built that can generate scripted dialogues that are
subsequently performed by animated human-like characters. One 
prototype to be delivered by {\sc neca} is an electronic showroom (eShowroom).\footnote{This application builds on the work carried out at {\sc Dfki} and reported in Andr\'{e} et al. (2000).} The idea is that a user/customer can select a class of cars and/or attributes (friendly for the environment, luxury, sportiness, etc.) in which s/he is interested. Furthermore, the user can set the personality traits of the characters which are to discuss this car (introverted, extroverted, agreeable, etc.). On the basis of these settings, the system then produces dialogues about specific cars and presents these to the user by means of embodied conversational agents which play out the dialogue. The strategies discussed in the present paper are highly 
relevant in the context of car sales. For example,
\begin{itemize}
\item Information about cars can be complex, so it can be
useful to have one or more participants ask
clarification questions (somewhat in the style of
Conan Doyle's Watson character, who triggers Sherlock
Holmes into explanations that benefit us as readers).
\item Not all customers are alike and it can be
useful to let different types of customers be represented
by different animated characters: one who is primarily
interested in the performance off the car, one who is
interested in chrome and gloss, one who is very aware of
environmental and safety-related issues, etc.
\end{itemize}
\noindent
Let us describe in more detail how one of the strategies of emphasis will be implemented in the {\sc neca} system. The {\sc neca} system generates the interaction between two or more characters in a number of steps, where information flows from a Scene Generator to a Multi-modal Natural Language Generator, to a Speech Synthesis component, to a Gesture Assignment component, and finally to a media player. 

In the Scene Generator, the basic structure of the dialogue is determined. For this purpose, a top-down planning algorithm is used. The output of this module is a {\sc rrl} Scene Description (see Piwek et al., 2002). Amongst other things, this Scene Description contains a set of dialogue acts. Individual dialogue acts are specified in terms of the dialogue act type, the speaker, the addressees, the semantic content, the actions which the act is a reaction to, and the and emotions (felt and expressed). The temporal ordering amongst the dialogue acts is represented separately and allows for underspecification. 

A Scene Description is constructed stepwise. We might start by constructing the following dialogue fragment:\footnote{In reality, at this stage in the processing linguistic realization has not yet taken place; thus the texts in E5. should be understood as mere paraphrases of the abstract descriptions of the dialogue acts which are actually passed on.}

{\small
\begin{tabbing}
xxxxx \= xxx \= xxx \= \kill
E5 \> $x_1$ \> B: \> How fast is this car? \\
\> $x_2$ \> S:  \> Its top speed is 180mph. \\
\> $x_2$ \> B:  \> Wow, that's great. 
\end{tabbing}    
}

\noindent
At this point, further elaboration of the dialogue is possible. 
Assume, for example, that the positive information that the car has a top speed of 180mph is to be emphasized. For this purpose, the strategy of emphasis we discussed in section \ref{emphasis} can be employed: a subdialogue can be inserted after $x_2$ consisting of a question by the buyer (`As much as 180mph?') which provides the seller with the opportunity to repeat a positive piece of information. This procedure yields the following `enhanced' dialogue:  

{\small
\begin{tabbing}
xxxxx \= xxx \= xxx \= \kill
E6 \> $x_1$ \> B: \> How fast is this car? \\
\> $x_2$ \> S:  \> Its top speed is 180mph. \\
\> $y_1$ \> B:  \> As much as 180mph? \\
\> $y_2$ \> S:  \> Yes, no less than 180 mph. \\
\> $x_2$ \> B:  \> Wow, that's great. 
\end{tabbing}    
}

\noindent
Note that the information which required emphasis has been mentioned no less than three times in the dialogue. This has been achieved by exploiting a very natural dialogue phenomenon: the occurrence of confirmation subdialogues. 

The thus created abstract representation of the dialogue (the Scene Description) can subsequently be processed further by the Multi-modal Natural Language Generator, the Speech Synthesis component, and the Gesture Assignment component. The result is sent to a media player which displays the dialogue to the user by means of a collection of embodied conversational characters.

\section{Conclusions}

We are not aware of any work which lays out 
strategies for the automated generation of scripted dialogue. There is a rapidly growing body of work on (Embodied) Conversational Agents (see, e.g., Ball \& Breeze, 1998; De Carolis et al., 2001; Loyall \& Bates, 1997; Nitta et al., 1997; Prendinger \& Ishizuka, 2001; Walker et al., 1996 and Zinn et al., 2002), but to the extent that language generation is discussed there\footnote{See Piwek (2002) for an overview of the literature in this area, specifically on affective natural language generation.}, it is from the perspective of the agents who participate in the conversation, rather than from the perspective of an author who produces a script for their interaction. The only exception we came across is Andr\'e et al. (2000) which reports on implemented systems for both spontaneous and scripted dialogue. Our aim has been to take a step back from the domain-specific implementation which they propose and find out whether it is possible to first identify more general strategies which are valid for the automated generation of scripted dialogue. We have tried to find such strategies on the basis of examples of human-authored and naturally occurring dialogues. We hope that our tentative investigations will encourage further studies into this new topic. A topic which, in our opinion, harbours interesting research questions at the intersection between dialogue processing and text generation, and which also lends itself well for various types of practical applications.

\section*{Acknowledgements}

We would like to thank Alexandre Direne, Martin Klesen and Neil Tipper for comments on an earlier draft of this paper, and two anonymous reviewers for their comments and encouragement.
This research is supported by the EC Project {\sc neca} IST-2000-28580. The information in this document is provided as is and no guarantee or warranty is given that the information is fit for any particular purpose. The user thereof uses the information at its sole risk and liability.

\section*{References}

\footnotesize{
\begin{description}

\item[] Andr\'{e}, E., T. Rist, S. van Mulken, M. Klesen \& S. Baldes (2000). `The Automated Design of Believable Dialogues for Animated Presentation Teams'. In: J. Cassell, J. Sullivan, S. Prevost and E. Churchill, {\em Embodied Conversational Agents}, MIT Press, 220-255. 

\item[]Ball, G. \& J. Breeze (1998).  `Emotion and
Personality in a Conversational Character'. In: {\it Proceedings of the
Workshop on Embodied Conversational Characters}, Lake Tahoe, CA, 1998,
83--86.

\item[]Cox, R., J. McKendree, R. Tobin, J. Lee \& T. Mayes (1999). `Vicarious learning from dialogue and discourse'. {\it Instructional Science}, 27, 431--458.

\item[]Copi, Irving M. (1972). {\em Introduction to Logic: Fourth Edition}. Macmillan, New York.

\item[] de Carolis, B., V. Carofiglio, C. Pelachaud \& I. Poggi (2001).
`Interactive Information Presentation by an Embodied Animated Agent'.
In: {\it Proceedings of the International Workshop on Information
Presentation and Natural Multimodal Dialogue}, Verona, Italy, 14--15
December 2001, 19--23.

\item[] Erasmus, Desiderius (1522). {\em Colloquia}.

\item[] Hovy, Eduard H. (1988). {\em Generating Natural Language Under Pragmatic Constraints}. Lawrence Erlbaum Associates, Hillsdale, New Jersey.

\item[]Lee, J., F. Dineen, J. McKendree (1998). `Supporting student discussions: it isn't just talk'. {\it Education and Information Technologies}, 3, 217--229.

\item[]Loyall, A.B. \& J. Bates (1997).  `Personality-Rich Believable Agents  That Use Language'. In: {\it Proceedings of the first International Conference on Autonomous Agents}, Marina Beach Marriott Hotel, Marina del Rey, California, February 5-8, 1997.

\item[]Nitta, K., O. Hasegawa, T. Akiba, T. Kamishima, T. Kurita, S. Hayamizu, K. Itoh, M. Ishizuka, H. Dohi, and M. Okamura (1997). `An Experimental Multimodal Disputation System'. In: {\it  Proc. of the
IJCAI-97 Workshop on Intelligent Multimodal Systems}, Nagoya, 1997.

\item[] Piwek, P. (2002). `An Annotated Bibliography of Affective Natural Language Generation'. {\it ITRI Technical Report ITRI-02-02}, University of Brighton. (available at: http://www.itri.bton.ac.uk/$\sim$Paul.Piwek)

\item[] Piwek, P., B. Krenn, M. Schr\"oder, M. Grice, S. Baumann \& H. Pirker (2002). `RRL: A Rich Representation Language for the Description of Agent Behaviour in NECA'. In: Proceedings of the AAMAS workshop ``Embodied conversational agents -- let's specify and evaluate them!'', 16 July 2002, Bologna, Italy.

\item[]Prendinger, H. \& M. Ishizuka (2001). `Agents That Talk Back
(Sometimes): Filter Programs for Affective Communication'. Contribution
to: {\it Second Workshop on Attitude, Personality and Emotions in
User-adapted Interaction} (in conjunction with User Modeling 2001),
Sonthofen, Germany, July 13, 2001.

\item[]Rickel, J., S. Marsella, J. Gratch, R. Hill, D. Traum \& W. Swartout (2002). `Toward a New Generation of Virtual Humans for Interactive Experiences'. {\it IEEE Intelligent Systems}, July/August 2002.

\item[]de Rosis, F. \& F. Grasso (2000).  `Affective
Natural Language Generation'. In: A.M. Paiva (Ed.), {\it Affective
Interactions}, Springer Lecture Notes in AI 1814, 204--218. 

\item[]Scott, D.C., B. Gholson, M. Ventura, A.C. Graesser and the Tutoring Research Group (2000). `Overhearing Dialogues and Monologues in Virtual Tutoring Sessions: Effects on Questioning and Vicarious Learning'. {\it International Journal of Artificial Intelligence in Education}, 11, 242--253. 

\item[]Searle, John R. (1983). {\em Intentionality}. Cambridge University Press, Cambridge.

\item[]Thomas, Jenny A. (1989). `Discourse control in confrontational interaction'.  In. L. Hickey (ed.), {\em The Pragmatics of Style}, Routledge, London and New York.

\item[]Twain, Mark (1917). {\em What is man? And other essays}. Chatto \& Windus, London.

\item[]Walker, M.A., J.E. Cahn \& S.J. Whittaker (1996).  `Linguistic
Style Improvisation for Lifelike Computer Characters'. In:  {\it
Proceedings of the AAAI Workshop on AI, Artificial Life and Entertainment},
Portland.

\item[] Zinn, Claus, Johanna D. Moore, \& Mark G. Core (2002). `A 3-tier Planning Architecture for Managing Tutorial Dialogue', To appear in: {\em Proceeding of Intelligent Tutoring Systems, Sixth International Conference} (ITS 2002), Biarritz, France, June 2002.

\end{description}
}

\end{document}